\documentclass{ecai}
\usepackage{graphicx}
\usepackage{latexsym}
\usepackage{amsmath}
\usepackage{xcolor}
\usepackage{multirow}

                  \newcommand{\PDTree}{\mbox{\sf SDebT}\/}

\newcommand{\UDG}{\mbox{\sf UDebG}}

\newtheorem{definition}{Definition}
\newtheorem{proposition}{Proposition}
\newtheorem{hypotesis}{Hypothesis}

\begin{document}

\begin{frontmatter}

\title{On the Complexity of the Bipartite Polarization Problem: from Neutral to Highly Polarized Discussions}

\author[A]{\fnms{Teresa}~\snm{Alsinet}
\thanks{Corresponding Author. Email: teresa.alsinet@udl.cat.}}
\author[A]{\fnms{Josep}~\snm{Argelich}}
\author[A]{\fnms{Ramón}~\snm{Béjar}} 
\author[A]{\fnms{Santi}~\snm{Martínez}} 

\address[A]{Departament d'Enginyeria Informàtica i Disseny Digital (DEIDD),  University of Lleida, Spain}


\begin{abstract}
The Bipartite Polarization Problem is an optimization problem
where the goal is to find the
highest polarized bipartition on a weighted and labelled graph that
represents a debate developed through some social network, where nodes
represent user’s opinions and edges agreement or disagreement between
users. This problem can be seen as a generalization of the maxcut problem,
 and in previous work approximate solutions and exact solutions have
 been obtained for real instances obtained from Reddit discussions, showing
 that such real instances seem to be very easy to solve.
In this paper, we investigate further the complexity of this problem, by introducing
an instance generation model where a single parameter controls the polarization
of the instances in such a way that this correlates with the average complexity
to solve those instances.
The average complexity results we obtain are consistent with our hypothesis:
the higher the polarization of the instance, the easier is to find the corresponding
polarized bipartition.
\end{abstract}

\end{frontmatter}

%
\section{Introduction}

The emergence of polarization in discussions on  social media networks, and the responsibility of companies in this problem, is a topic that is causing a significant interest among society~\cite{Lorenz-Spreen_2023}, although the complete relationship between digital platforms and polarized attitudes remains unclear~\cite{barbera_2020}. 
For example, there is some work that indicates that, at least for
polarization in people's views of political parties,
the increased use of social networks and the internet may not be necessarily increasing
polarization~\cite{Boxell20}.

Online social networks are platforms for social and communicative interaction. 
Studies have shown that polarization varies greatly across platforms~\cite{Yarchi_2021}, the strength of the results being a function of how polarization is measured~\cite{Kubin2021}. 
Previous work has studied the presence of polarization in different concrete examples, trying to analyze the relevant characteristics in these cases. For~example, the~works~\cite{echoPLOS,echoFacebook} studied the emergence of so-called echo chambers, where users are found to interact mainly only with the users that agree with them and to have very few interactions with other groups.
However, online discussions in social networks can also show polarization where there are answers with negative sentiment between different groups, which can be considered the most worrying situation. For~example, in~\cite{DBLP:conf/icwsm/RecueroSG20} they studied 
hyperpartisanship and polarization in Twitter during the 2018 Brazilian
presidential election. Their findings showed that there was more interaction within each group (pro/anti Bolzonaro) but there was also an interaction between both groups.
Actually, there are also cases where the interaction between groups can be more relevant than those within groups, like in the case studied in~\cite{noechoReddit} where the analysis of the 2016 U.S. elections on Reddit showed a significant number of interactions between pro-Trump and pro-Hillary supporters.
So, the extent to which bias is due to inline echo chambers and filter bubbles is also misunderstood, with evidence pointing in opposite directions~\cite{barbera_2020}.
A major problem is that divisive content tends to spread widely and quickly on social media. Posts that express moral outrage~\cite{Brady_2017} or bash one's outparty~\cite{Rathje_2021}, for example, tend to be particularly successful at going viral. This virality is driven by the prioritizations made by social media algorithms coupled with people's general inclination to favor sensational content. As a consequence, when people log on to social media, they are likely to see content that is divisive and presses their emotional buttons. In addition, these trends encourage politicians, media outlets and potential influencers to publish divisive content because it is most likely to give the commitment they desire.
To try to mitigate the factors that may be helping the spread of divisive content, for instance, 
Facebook has  launched some initiatives~\cite{FBblog1}, even if this kind of content may be the one that produces
the maximum attention of their users, so being also the one producing maximum economic benefit.

Because each online social  network company can have its own personal interest regarding when to control this kind of behaviour,  one fundamental aspect is to define more transparent ways to monitor
such possible non-desirable behaviours, so that we can 
inform solutions for creating healthier social media environments.
From our point of view, the goal to be achieved in this direction is to provide  mechanisms that allow us to detect situations where
one can deduce that polarization is taking place, and to a certain level of severity, because 
there is some objective value we can measure for this. 

With this goal in mind and given that 
Reddit turned out to be a decisive political debate platform in different elections in the United States, such as those of 2016, a research work introduced a problem~\cite{app112411879} about finding the highest polarized
bipartition on a weighted and labelled graph that represents a Reddit debate, where nodes represent user's opinions and edges 
agreement or disagreement between users. 
The authors showed that
finding this target bipartition is an optimization problem that can be seen as a generalization of the maxcut problem, and they introduced a basic local search
algorithm to find approximate solutions of the problem. Later on, in~\cite{DBLP:conf/ccia/AlsinetABM22},  the complexity of solving real 
instances from Reddit discussions with both complete and approximate algorithms was also studied. These results on real Reddit instances showed that solving those instances could be performed 
with little computational cost as the size of the instances increased, and that all the instances, obtained from different subreddits, tended to show polarization values away from neutral discussions, although without reaching the case of extreme polarization.

Our aim in this paper is to further investigate
the complexity of this problem,
by introducing an instance generation model where a single parameter controls the polarization of the instances in such a way that this correlates with the average complexity to solve those instances. The average complexity results we obtain are consistent with our hypothesis: the higher the polarization of the instance, the easier is to find the corresponding polarized bipartition.
Although one can consider other alternative ways to measure the polarization in online social networks, we believe that our generation model could be used to easily  check  other measures.
The results obtained with our generation model together with the previous results on real instances from Reddit
discusions  seem to indicate that in practice, it may be feasible to compute the polarization of online discussions, at least with a polarization measure like the one we use here.

Therefore, in view of the experimental results,  checking polarization with this measure seems to be difficult only in unrealistic cases.
Clearly, it is advantageous to can offer algorithms for monitoring polarization
in online social networks.



The structure of the rest of the paper is as follows. In~Section~\ref{sec:problem}, we
 present both the model for 
 weighted and labelled graphs that represents an online debate
 and the measure to quantify the polarization in such debate graphs,  studied and developed in~\cite{app112411879}. In Section~\ref{sec:nphardness} we prove that the computation of such measure
 is NP-hard.
In Section~\ref{sec:algorithms} we describe the algorithms that we use to solve the Bipartite Polarization Problem.
In Section~\ref{sec:randomgen},  we {introduce}  our random generation model for user debate graphs based on a main parameter, called $\alpha$, to control the polarization of the instances.
Finally, in~Section~\ref{sec:experiments}, we perform an empirical evaluation to study 
how the complexity of solving the typical instances obtained with our random model 
changes as we modify  the parameter $\alpha$ and how this at the same time affects 
to the polarization of the  instances.


\section{Problem Definition}
\label{sec:problem}


Polarization corresponds to the social process whereby a  group of persons is divided into two opposing subgroups having conflicting and contrasting positions, goals and viewpoints, with few individuals remaining neutral or holding an intermediate position~\cite{Sunstein_2002}.
The model we consider in this work is inspired by the model used in~\cite{DBLP:conf/coling/MurakamiR10} to identify supporting or opposing opinions in online debates, based on finding a maximum cut in a graph that models the interactions between users.
Moreover, 
as~we are interested in quantifying polarization, we consider a model
that is based on a weighted graph and with labeled edges, where node weights represent the side of the user in the debate and edge labels represent the overall sentiment between two users. 
Then, given a bipartition of this graph, the polarization degree of the bipartition is based on how homogeneous each partition is and how negative the interactions are between both partitions.   
Finally, the measure of debate polarization is based on the maximum polarization 
obtained in all the possible bipartitions of the~graph.

Adapting from~\cite{app112411879}, an {\em online debate} $\Gamma$ on a root comment $r$ is a non-empty set of comments, that were
originated as successive answers to a root comment $r$. 
An online debate is modeled as a two-sided debate tree, 
where nodes are labelled with a binary value  that denotes whether the comment is in agreement (1)  or in disagreement
(-1) with the root comment.~\footnote{In~\cite{app112411879} they introduced this model specifically for Reddit debates.
However, it is clearly a model suitable for other similar social networks.}

\begin{definition}[Two-Sided Debate Tree] 
  Let $\Gamma$ be an online debate on a root comment r.
A  {\em Two-Sided Debate Tree} {\sf (\PDTree)} for $ \Gamma$
%
 is a tuple  $ \mathcal{T}_S = \langle  C, r, E, W , S \rangle $ defined  as~follows:


\begin{itemize}
\item For every comment $c_i$ in $\Gamma$, there is a node $c_i$ in $C$.
\item Node $r \in C$ is the root node of $\mathcal{T}$.
\item If a comment $c_1 \in C$ answers  another comment $c_2\in C$,  there is a directed edge $(c_1, c_2)$ in $E$.
\item $W$ is a labelling function  of answers  (edges)
$W : E \rightarrow [-2,2]$, where the value assigned
to an edge $(c_1, c_2) \in E$
denotes the  sentiment of the answer  $c_1$ with respect to $c_2$, from~highly negative ($-2$) to highly positive ($2$).
\item $S$ is a labelling function  of comments (nodes)
$S : C \rightarrow \{-1, 1\}$, where  the value assigned
to a node $c_i \in C$
denotes whether the comment $c_i$ is in agreement (1)  or in disagreement
(-1) with the root comment $r$ and it is defined as follows:

\begin{itemize}

\item[-]  $S(r) = 1$ and


\item[-] For all node $c_{1} \neq r$ in  $C$, $S(c_{1}) = 1$ if for some node $c_{2} \in C$, $(c_{1},c_{2}) \in E$ and either $S(c_{2}) = 1$ and
$W(c_{1},c_{2}) > 0$, or~$S(c_{2}) = -1$ and
$W(c_{1},c_{2}) \leq 0$;  otherwise, $S(c_{1}) = - 1$.
\end{itemize}
%
\end{itemize}
Only the nodes and edges obtained by applying this  process belong to $C$ and $E$,
 respectively.

\end{definition}

Next, we present the formalization of a User Debate Graph based on a
 Two-Sided Debate Tree, where now  all the comments of a same user are 
 aggregated
 into a single node that represents the user's opinion.

\begin{definition}[User Debate Graph]

Let $\Gamma$ be a online  debate on a root comment $r$
 with users' identifiers  $U=\{u_1, \ldots, u_m\}$ and let
 $ \mathcal{T}_S =   \langle  C, r, E, W , S \rangle $ be a  \PDTree\    for $\Gamma$.
A  {\em User Debate Graph\/}  {\mbox{\sf (\UDG)}\/} for
$\mathcal{T}_S$
is a tuple $\mathcal{G} =\langle {\mathcal{C}},  {\mathcal{E}}, {\mathcal{S}}, {\mathcal{W}} \rangle$, where:
 \begin{itemize}
\item $\mathcal{C}$ is the set of nodes  of $\mathcal{G}$ 
defined as
the set of
 users' opinions $\{C_{1}, \ldots, C_{m}\}$;
 i.e.,~$\mathcal{C} = \{C_{1}, \ldots, C_{m}\}$
 with
 $C_{i}=\{ c \in \Gamma \mid  c \neq r {\textrm{ and user}(c) = u_i}\}$, for~all users $u_i \in U$.

%
%


\item $\mathcal{E} \subseteq \mathcal{C} \times \mathcal{C}$ 
is the set of edges of $\mathcal{G}$ defined as the set of
interactions between different users in the debate;
i.e.,
there is an edge $(C_{i}, C_{j})\in \mathcal{E}$, with~$C_{i}, C_{j}  \in \mathcal{C}$ and $i \neq j$, if~and only if
for some  $(c_1,c_2) \in E$ we have that
$c_1 \in C_{i}$ and $c_2 \in C_{j}$.

\item $\mathcal{S}$ is an opinion weighting
scheme for $\mathcal{C}$ that expresses the side of users in the debate based on the side of their comments.
We define
$\mathcal{S}$ as
the mapping
$\mathcal{S}: \mathcal{C} \rightarrow [-1, 1]$
that assigns to every node $C_{i} \in \mathcal{C}$
  the  value
 $$ \mathcal{S}(C_{i}) = \frac{\sum_{c \in C_{i}} S(c_{i})}{|C_{i}|}$$
  in the real interval $[-1, 1]$ that expresses the side of the user $u_i$  
 with respect to the root comment, from~strictly disagreement ($-$1) to strictly agreement (1), going through undecided  opinions (0).

\item $\mathcal{W}$ is an interaction weighting  scheme for $\mathcal{E}$
 that expresses 
 the overall sentiment 
between users
  by combining the individual  sentiment values assigned  to the responses between their comments.
 %

%
We define
$\mathcal{W}$ as the mapping  $\mathcal{W}: \mathcal{E} \rightarrow  [-2,2]$
 that assigns to every edge $(C_{i}, C_{j}) \in \mathcal{E}$
 a  value
  $w  \in    [-2,2] $
  defined as follows:
%
%
$$w = \sum_{\{(c_1,c_2) \in E \cap (C_{i} \times C_{j}) \}} W(c_1, c_2) / |\{(c_1,c_2) \in E \cap (C_{i} \times C_{j})\}|$$
 where 
 $w$ expresses the
   overall sentiment of the user $u_i$  regarding the comments of the user  $u_j$, from~highly negative ($-$2) to highly positive (2).

\end{itemize}

Only the nodes and edges obtained by applying this  process belong to $\mathcal{C}$ and $\mathcal{E}$, respectively.
\end{definition}

Given a User Debate Graph  $\mathcal{G} =\langle {\mathcal{C}},  {\mathcal{E}}, {\mathcal{S}}, {\mathcal{W}} \rangle$,  a model to measure the level of polarization in the debate between its users was also introduced in~\cite{app112411879}. It is based on two characteristics that a polarization measure should capture.
First, a~polarized debate should contain a bipartition of ${\mathcal{C}}$ into two sets $(L,R)$ such that the set $L$ contains  mainly {users in disagreement}, the~set $R$ contains mainly {users in agreement}, and~both sets should be similar in size.
The second ingredient is the sentiment between users of $L$ and  $R$.
A polarized discussion should contain most of the negative interactions between users of $L$ and users
of $R$, whereas the positive interactions, if~any, should be mainly within the users of $L$ and within the users \mbox{of $R$.}

To capture these two characteristics with a single value, two different measures are combined
in a final one, referred to as {\em the bipartite polarization}.

\begin{definition}[Bipartite Polarization]
Given a User Debate Graph  $\mathcal{G} =\langle {\mathcal{C}},  {\mathcal{E}}, {\mathcal{S}}, {\mathcal{W}} \rangle$   and
a bipartition $(L,R)$ of ${\mathcal{C}} $, we define:
\begin{itemize}
\item The  level of consistency and  balance of $(L,R)$ is a real value in $[0,0.25] $ defined as follows:
 $$ {\textrm{\em SC}}(L,R,\mathcal{G}) =  {\textrm{\em LC}}(L,\mathcal{G}) \cdot  {\textrm{\em RC}}(R,\mathcal{G})$$
 with:
 $$ {\textrm{\em LC}}(L,\mathcal{G}) = \frac{\sum_{\substack{C_i \in L,\\ \mathcal{S}(C_i) \leq 0  }} -{\mathcal{S}}(C_i)      }
             { |{\mathcal{C}}| }  $$
and
 $$ {\textrm{\em RC}}(R,\mathcal{G}) =  \frac{\sum_{\substack{C_i \in R,\\ \mathcal{S}(C_i) > 0  }} {\mathcal{S}}(C_i)      }
             { |{\mathcal{C}}| }.  $$

\item
The~sentiment of the interactions between users of different sides is a real value in
 $[0, 4]$
defined as follows~\footnote{According with~\cite{DBLP:conf/ccia/AlsinetABM22}, each edge label $(i, j)$ can incorporate a correction factor that is used to modify the final weight used in this function. However, to simplify the model notation, we will consider that the weight $\mathcal{W}(i,j)$ already  reflects this factor.}:


$${\textrm{\em SWeight}}(L,R,\mathcal{G}) =  \frac{\sum_{\substack{(i,j) \in  {\mathcal{E}} \cap \\  ((L \times R) \cup (R \times L))}}  -\mathcal{W}(i,j) }
{ |{\mathcal{E}}|  }  + 2,$$


\end{itemize}

Then, the Bipartite Polarization  of $\mathcal{G}  $ on a bipartition $(L,R)$ is the value in the real interval $[ 0,1 ]$ defined as follows:

$$ {\textit{BipPol}}(L,R,\mathcal{G}) =  SC(L,R,\mathcal{G}) \cdot  {\textit{SWeight}}(L,R,\mathcal{G}).$$

%
Finally, the  Bipartite Polarization  of $\mathcal{G}  $
is the maximum value of $  {\textit{BipPol}}(L,R,\mathcal{G}) $
among all the possible bipartitions $(L,R)$.

\end{definition}

\section{Worst-case Complexity of the Bipartite Polarization Problem}\label{sec:nphardness}


\begin{proposition} 
The \UDG \ bipartite polarization problem is NP-hard.
\end{proposition}

Proof.
We prove that the simple maxcut problem, that it is NP-hard even for graphs with bounded degree $\leq 3$ ~\cite{DBLP:conf/stoc/Yannakakis78}, can be reduced to the \UDG \  bipartite polarization problem in polynomial time.
Consider an undirected graph instance $G=(V,E)$ of the simple maxcut problem.
Then, we build an instance $\mathcal{G} =\langle {\mathcal{C}},  {\mathcal{E}}, {\mathcal{S}}, {\mathcal{W}} \rangle $ of the bipartite polarization problem such that:
\begin{enumerate}
    \item  The set of vertices  ${\mathcal{C}}$ is equal to $ V \cup \{ u^-, u^+  \} $.
    For ${\mathcal{S}} $ we have that $ \forall v \in V , {\mathcal{S}}(v) = 0$ and 
    $ {\mathcal{S}}(u^-) = -1, {\mathcal{S}}(u^+) =  +1$.
     \item  The set of edges $ {\mathcal{E}} $ is defined only for those vertices that have an edge 
     in the input graph: $  {\mathcal{E}} = \{ (v_1,v_2), (v_2,v_1) \ | \ \{ v_1,v_2\} \in E \}  $.
     And for any $ e \in  {\mathcal{E}} $ we have that $  {\mathcal{W}}(e) = -1/2 $.
\end{enumerate}

Then, assume $G=(V,E)$ has a bipartition  $(L,R)$ with total weight (number of edges between $L$ and $R$) equal to $W$. Consider the bipartition $(L \cup \{ u^- \}, R \cup \{ u^+\})$ for the instance  $\mathcal{G}$ obtained by the reduction.
Clearly:
$$  {\textit{BipPol}}(L \cup \{ u^- \}, R \cup \{ u^+\},\mathcal{G}) ) = \frac{-(-1)}{| {\mathcal{C}}|} 
   \frac{1}{| {\mathcal{C}}|} \left(\frac{W}{|E|}+2\right)    $$

Next, assume $\mathcal{G} =\langle {\mathcal{C}},  {\mathcal{E}}, {\mathcal{S}}, {\mathcal{W}} \rangle $ has 
a bipartition $(L',R')$ with bipartite polarization value  equal to $BPol$.
Then, as all the nodes have $ {\mathcal{S}}(v) = 0$, except for  $ {\mathcal{S}}(u^-) = -1, {\mathcal{S}}(u^+) =  +1$ , we have that if $BipPol > 0$ then $u^-$ will be in $L'$ and 
$u^+$ will be in $R'$, so the BipPol value will be of the same form as before:

$$  {\textit{BipPol}}(L' , R' ,\mathcal{G}) ) = \frac{-(-1)}{| {\mathcal{C}}|} 
   \frac{1}{| {\mathcal{C}}|} \left(\frac{W}{|E|}+2\right)    $$
where $W$ will be equal to half the number of edges between $L'$ and $R'$ in $ \mathcal{G}$,
as  for any undirected edge in the  input graph $G$ we have two directed edges in
$ \mathcal{G}$ each one with $W(e)= -1/2$. 
But this implies that $ (L'\setminus \{ u^- \}, R'\setminus\{ u^+ \} )$ is a bipartition for  the input graph $G$ with value $W$.
For the case $BPol=0$, this could only happen because the vertices $u^-$ or $u^+$ are in
the wrong side.
But we can always transform any bipartition $(L',R')$ to one in which $u^-$ and $u^+$
are in the right side, and so derive the value of $W$ from the value of $BPol$.

As a consequence,   $(L,R)$ is a simple maximum cut  for $G=(V,E) $ if and only if $ (L \cup \{ u^- \}, R \cup \{ u^+\}) $ is 
a bipartition  for  $\mathcal{G} =\langle {\mathcal{C}},  {\mathcal{E}}, {\mathcal{S}}, {\mathcal{W}} \rangle $ with maximum  $ {\textit{BipPol}} $ value.

This reduction shows two interesting facts about our problem. First, 
we can have instances that are as hard to solve as the Maxcut 
problem when {\it almost} all the users have $  {\mathcal{S}}(v) = 0 $. But at the same time, observe that we need to have at least one 
user with negative value and another one with positive value. This 
is because if strictly {\it all} the users have $  {\mathcal{S}}(v) = 0 $ then the $ BipPol(L,R,{\mathcal G})$ value is equal to zero for any bipartition $(L,R)$ and then it becomes trivial to solve the problem.
These facts together with the experimental results 
in~\cite{DBLP:conf/ccia/AlsinetABM22} with real instances from Reddit discussions, where instances with polarization values away from neutral were very easy
to solve on average, makes us to present the following hypothesis.

\begin{hypotesis}
  On one hand,
  the closer the maximum bipartite polarization of a \UDG \ instance is to zero, the more difficult is to find such bipartition as 
  many possible bipartitions will have a bipartite polarization 
  very close to the optimum.
  On the other hand, the closer the maximum bipartite polarization of a \UDG \ instance is to one, the easier is to find such bipartition as any user $i$ will tend to have values for $ {\mathcal{S}}(i)$ 
  and $ \mathcal{W}(i,j) $ that are correlated, and users will tend
  to have only negative sentiment answers to the users of the other 
  side in the optimal bipartition.
\end{hypotesis}

The random generator model we present in Section~\ref{sec:randomgen} will be used in the experimental section to try to check this hypothesis.

\section{Solving the Bipartite Polarization Problem}\label{sec:algorithms}

To solve the Bipartite Polarization Problem we use two approaches, one based on a complete algorithm to find the exact polarization, and another one based on a local search algorithm that finds an approximate solution. The complete algorithm uses the same SCIP branch-and-bound solver (version 8.0)~\cite{BestuzhevaBesanconChenetal.2021} and the same integer nonlinear programming formulation (MINLP) used in~\cite{DBLP:conf/ccia/AlsinetABM22}. For the approximate approach, we use a local search solver inspired by an algorithm for the maxcut problem~\cite{BenlicH13}. The algorithm uses a greedy approach based on a steepest ascent hill climbing strategy plus restarts to escape from local minimas. The number of steps of the local search algorithm is bounded by the number of nodes, and the number of restarts is set to 10. So the worst case running time of the local search approach is linear in the number of nodes.

\section{A Random Generator Model for \UDG \ Instances}\label{sec:randomgen}

We present a random generator of  \UDG \ instances where the goal is to control the
expected bipartite polarization of the instance by means of a single parameter $\alpha$
that lies in the range $(0,1]$. The number of nodes (users) is given by a parameter $m$.
The generation process consists of the two following steps:
\begin{enumerate}
\item Generation of the set of nodes $ \mathcal{C} $ with their $\mathcal{S}(C_{i}) $ value.
Each node $C_i$ is generated with a value  $\mathcal{S}(C_{i}) $ from $ [-\alpha,\alpha ]$ obtained with a bimodal distribution  that arises as a mixture of two different truncated normal distributions $TN(-\alpha,0,\mu_1,\sigma_1)$
  and $TN(0,\alpha,\mu_2,\sigma_2)$
\footnote{Note that for $\alpha=0$,
all the nodes would have $\mathcal{S}(C_{i}) =0$. 
This extreme case corresponds to the trivial problem where the bipartite polarization is zero for any bipartition.
For this reason, $\alpha=0$ is not considered in our random generator.}.
The first distribution is defined on the interval $[-\alpha,0]$ with mean $\mu_1$ and standard deviation $\sigma_1$
equal to:
$$ \mu_1=-\alpha, \ \sigma_1 =\frac{1}{1+20\alpha}$$
and anologously for the second distribution, but now defined on the interval $ [ 0,\alpha ]$ and with values:
$$ \mu_2=\alpha,\ \sigma_2 = \frac{1}{1+20\alpha}$$
So, with this bimodal distribution the values are concentrated around the two modes ($-\alpha$ and $\alpha$), but
how tightly they concentrate is inversely proportional to $\alpha$, so the higher the value of
$\alpha$ the smaller the standard deviation.


%
\item Generation of the set of edges $\mathcal{E}$ with their $\mathcal{W}(i,j)$ value.
For each node $i$, we randomly select a set of $k$  target vertices $\{j_1,j_2,\ldots,j_k \}$,
with $k$ randomly selected from $[1,\lceil \log_{10}(m) \rceil]$, to
build outgoing edges from $i$ to those vertices. The value of $ \mathcal{W}(i,j) $ is generated
with a truncated normal distribution on the interval $[-2,2]$ with $\mu$ and $\sigma$ that depend on the values of $\mathcal{S}(C_{i}) $ and $ \mathcal{S}(C_{j})$ as follows:

$$ \mu = \left\{ \begin{array}{ll}
 2 \cdot |\mathcal{S}(C_{i})|-| \mathcal{S}(C_{i})- \mathcal{S}(C_{j})| & \textrm{if} \  C_{i},C_{j} \textrm{\  \ same side}\\
 -|\mathcal{S}(C_{i})| \cdot | \mathcal{S}(C_{i})- \mathcal{S}(C_{j})| & \textrm{if} \  C_{i},C_{j} \textrm{\  \ different side}\\
\end{array} \right. $$
$$ \sigma =  \frac{2}{3 + 10|\mu|} $$
So,  when the users $C_{i}$ and $C_{j}$ are on the same side ($\mathcal{S}(C_{i})$ and $ \mathcal{S}(C_{j})$ are both either positive or $\leq 0$), the mean of the distribution will be positive and
the more similar are the values  $\mathcal{S}(C_{i}) $ and $\mathcal{S}(C_{j}) $  and the closer is  $|\mathcal{S}(C_{i})| $ to 1, the closer will be $\mu$ to 2.
By contrast, when the users $C_{i}$ and $C_{j}$ are on different sides,
the more different are the values $\mathcal{S}(C_{i}) $ and $\mathcal{S}(C_{j}) $ and the closer is
$|\mathcal{S}(C_{i})| $ to 1 , the
closer will be $\mu$ to  -2. 
So, observe that the sign of $\mu$ depends on the sign of users $i$ and $j$. Regarding the absolute value $|\mu|$, it depends
on both  $ \mathcal{S}(C_{i})  $ and  $ \mathcal{S}(C_{j})  $, but more on the first one. This is because in principle the overall
sentiment $ \mathcal{W}(i,j) $ (answers from user $i$ to user $j$) is not necessarily equal to  $ \mathcal{W}(j,i) $, although it is natural to think that there will be some positive correlation between them.
In any case, we think that 
it makes sense to force the sign of the mean to be the same
one when generating $ \mathcal{W}(i,j) $ and $ \mathcal{W}(j,i) $.
As for the absolute value, it is natural that users with stronger opinions (those with $|\mathcal{S}(C_{i})|$ closer to $1$) have stronger sentiments towards the rest of the users.

\end{enumerate}

The overall intention of this way to select the values of $\mathcal{S}(C_{i})  $ and $\mathcal{W}(i,j) $ is to move from 
a \UDG \ instance corresponding to a clearly neutral discussion 
when $\alpha$ approaches $0$ to one corresponding to a highly polarized discussion when $\alpha$ approaches $1$.
We control this by making the expected values of both $\mathcal{S}(C_{i})$ and $\mathcal{W}(i,j)$ to approach the neutral value $0$ as $\alpha$ approaches $0$.
Observe, however, that even
in the case when $\mu=0$ for the generation of $\mathcal{W}(i,j)$
the standard deviation is $2/3$.
This allows the generation of some negative and positive values around the mean
to ensure a more realistic generation of users where 
not all the answers to any user are strictly neutral.
Then, as $\alpha$ approaches $1$, the expected value for 
 $\mathcal{S}(C_{i}) $ tends to be more tightly concentrated 
 around -1 or 1 due to the standard deviation tending to 
 $ 1/21$ and 
 the expected value for $ \mathcal{W}(i,j) $ is more concentrated 
 around the extreme values $-2$ and $2$ due to the standard 
 deviation tending to $2/13$. 
 
 We make that for $\alpha=1$ the standard deviation for the 
 generation of $\mathcal{S}(C_{i}) $  and $ \mathcal{W}(i,j) $
 to not be $0$, but very low, because we do not believe that in
 practice one encounters real discussions with such extreme polarization value of 1. So, we have preferred instead to move 
 towards a zone of high polarization, but without reaching the
 extreme case. However, it is very easy to expand the range of 
 possible polarizations by simply using other values to move 
 the standard deviation of both distributions closer to 0
 as $\alpha$ approaches $1$.

To select the number of out edges for each node, we use the function $\lceil log_{10}(m) \rceil$ as the maximum number of out edges per node. Once more, we use this function as we believe this represents better the number of interactions of real debates in social networks. If we compute the maximum out degree and mean out degree of each instance used in~\cite{DBLP:conf/ccia/AlsinetABM22}~\footnote{Instances obtained from the authors of the paper.}, we obtain that the mean of the maximum out degree of all the instances is 4.05, and the mean of the mean out degree of all the instances is 0.89, where the instances have a median number of nodes around 50~\footnote{There are some nodes (users) that only answer to the root node. As the answers to the root node are not considered for the \UDG, they do not have out edges, so it is possible to have a mean out degree below 1 in the \UDG.}. With those numbers in mind, the $log_{10}(m)$ function allows us to limit the maximum out degree to a realistic number of edges as   $m$ increases, and at the same time keep a low mean value.


\begin{figure*}[htb]
    \centering
    \includegraphics[width=0.49\textwidth]{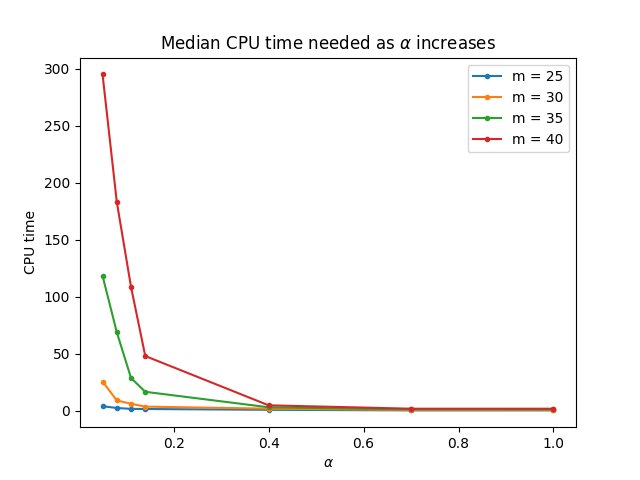}
    \includegraphics[width=0.49\textwidth]{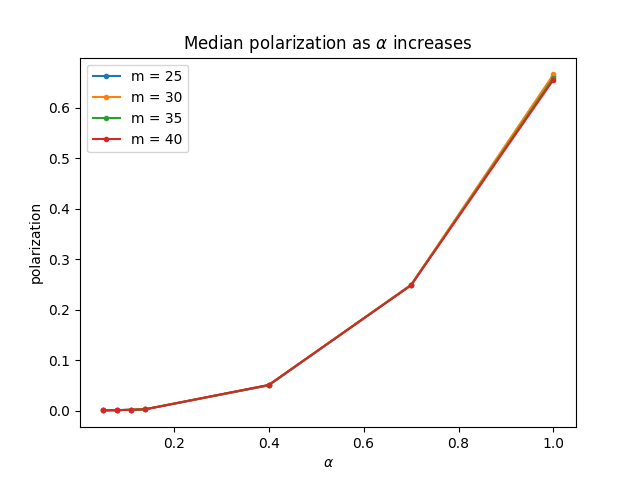}
    \caption{CPU time needed to solve the instances (left plot) and polarization of the solution (right plot) as we increase the alpha value for instances with nodes ranging from 25 to 40.}
    \label{fig:time}
\end{figure*}

\section{Experimental Results}\label{sec:experiments}

In this section we present an empirical evaluation of the complexity to solve the instances obtained with 
our random generator for different values of the parameter $\alpha$ and increasing the number of nodes (users) of the
instances. Our main interest is to understand how the complexity of the exact algorithm based on integer programming
changes as $\alpha$ moves from almost $0$ to $1$. As a second goal, we want also to compare the quality of the approximate 
polarization value returned by the algorithm based on local search. The results reported in~\cite{DBLP:conf/ccia/AlsinetABM22} indicate that for real instances, coming from Reddit discussions, the  median time complexity was always very low and that the approximation of the 
polarization value given by the local search algorithm was always almost identical to the 
exact value.
The experiments were performed on an Intel\textregistered~Core\textregistered~i5-6400 CPU at 2.70GHz with 16GB of RAM memory.

The results of the median time to solve instances with $\alpha \in \{0.05,0.08,0.11,0.14,0.4,0.7,1.0 \}$
 and $m \in \{25,30,35,40 \}$ as well as the bipartite polarization of the instances are shown on 
 Figure~\ref{fig:time}. For each combination of $\alpha$ and $m$, the median shown on the figure is 
 obtained from a set of 50 instances obtained with our random generator.
 We have used close values of $\alpha$ up to $0.14$ because as one can observe, the median time to solve the instances increases quickly as $m$ increases for very low values of $\alpha$, but as we move away from $0$ the median time decreases abruptly.
 At the same time, the median bipartite polarization of the instances increases slowly for low values of $\alpha$ but then it starts to increase more quickly as $\alpha$ increases.
So, these results are consistent with our hypothesis about the relation between polarization and complexity to solve the instances.

\begin{figure}[htb]
    \includegraphics[width=0.49\textwidth]{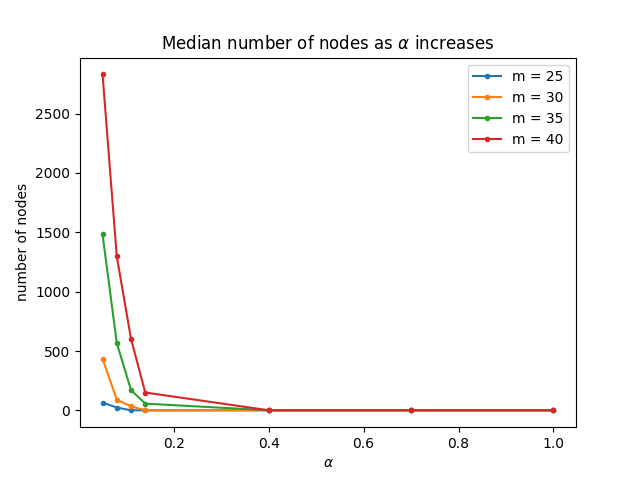}    
    \caption{Number of nodes of the SCIP search tree needed to solve the instances as we increase the alpha value for instances with nodes ranging from 25 to 40.}
    \label{fig:nodes}
\end{figure}

 If we compare these results with the ones with real instances from Reddit in~\cite{DBLP:conf/ccia/AlsinetABM22},
 that have median number of nodes around 50, we observe that they are also consistent, in the sense 
 that in the Reddit discussions used in that paper the median polarization of the instances was around 0.5,  and the median time was around 2 seconds. To make a more fair comparison of the complexity, given that we do not know the exact hardware used in that paper to solve the instances, we have also computed the median number of nodes performed by the exact algorithm based on the SCIP solver (the same one used in that previous paper), given that in the previous
 paper the authors also showed the median number of nodes of the SCIP search tree. 
 Figure~\ref{fig:nodes} shows the median number of nodes for the same instances 
 of Figure~\ref{fig:time}. As we can observe, the number of nodes follows the same qualitative behavior as the time, 
 and for instances with median polarization around 0.5 (that have $\alpha > 0.7$) the median number of nodes would 
 be around 1 (by interpolating between our cases $\alpha=0.7$ and $\alpha=1.0$), like it happens in the results 
 of~\cite{DBLP:conf/ccia/AlsinetABM22} with the Reddit instances.

Finally, we were also interested in comparing the quality of the approximation provided by the local search approach, presented 
in Section~\ref{sec:algorithms}, with the one of the exact algorithm. Table~\ref{table1} shows, for each combination of number 
of nodes and value of $\alpha$, the minimum, median and maximum value  for the polarization obtained in each set of instances 
with the exact algorithm (columns labeled with {\it SCIP BipPol}).
Observe that is for higher values of $\alpha$ where we observe more variations in the polarization of the instances (difference between the min and the max values). Then, in the next three columns we show for the same three instances of the previous columns (instance with minimum polarization, with median one, and with maximum one) the ratio of the approximate polarization computed by the local search solver to the exact polarization of the instance (LS ratio). As we observe, the quality of the solutions computed by the local search algorithm is perfect for instances with $\alpha \geq 0.4$ (that are precisely the ones that are easily solved by the exact algorithm) and for lower values of $\alpha$ we observe only a tiny relative error.
Overall, we can say that the quality of the solution provided by the local search algorithm is always satisfactory with the 
advantage that its computational complexity is always linear with respect to the number of nodes, so there is no hard region for the local search approach.

\begin{table}
\begin{center}
{\caption{Comparison of the quality of the solutions between SCIP and LS solvers for different $\alpha$ values and number of nodes in the range 25-40.}\label{table1}}
\begin{tabular}{c|c|ccc|ccc}
 & & \multicolumn{3}{c|}{SCIP BipPol} & \multicolumn{3}{c}{LS ratio} \\
$m$ & $\alpha$ & min & median & max & min & median & max \\
\hline
\multirow{7}{*}{25} & 0.05 & 0.0001 & 0.0003 & 0.0005 & 1 & 0.998 & 1 \\
 & 0.08 & 0.0004 & 0.0008 & 0.0013 & 1 & 0.9995 & 0.9999 \\
 & 0.11 & 0.0007 & 0.0015 & 0.0025 & 1 & 0.9998 & 1 \\
 & 0.14 & 0.0013 & 0.0025 & 0.0042 & 1 & 1 & 0.9999 \\
 & 0.4 & 0.0359 & 0.0497 & 0.0611 & 1 & 1 & 1 \\
 & 0.7 & 0.1907 & 0.2487 & 0.2786 & 1 & 1 & 1 \\
 & 1.0 & 0.4834 & 0.6631 & 0.7446 & 1 & 1 & 1 \\
\hline
\multirow{7}{*}{30} & 0.05 & 0.0002 & 0.0003 & 0.0005 & 0.9963 & 0.9985 & 0.9989 \\
 & 0.08 & 0.0005 & 0.0008 & 0.0012 & 0.9989 & 0.9995 & 0.9992 \\
 & 0.11 & 0.0010 & 0.0015 & 0.0023 & 0.9997 & 0.9997 & 1 \\
 & 0.14 & 0.0017 & 0.0025 & 0.0038 & 0.9998 & 0.9999 & 0.9999 \\
 & 0.4 & 0.0424 & 0.0505 & 0.0597 & 1 & 1 & 1 \\
 & 0.7 & 0.2046 & 0.2492 & 0.2724 & 1 & 1 & 1 \\
 & 1.0 & 0.5289 & 0.6657 & 0.7459 & 1 & 1 & 1 \\
\hline
\multirow{7}{*}{35} & 0.05 & 0.0002 & 0.0003 & 0.0005 & 1 & 0.9991 & 1 \\
 & 0.08 & 0.0006 & 0.0008 & 0.0012 & 1 & 0.9996 & 1 \\
 & 0.11 & 0.0011 & 0.0015 & 0.0022 & 0.9996 & 0.9999 & 1 \\
 & 0.14 & 0.0018 & 0.0025 & 0.0037 & 0.9998 & 1 & 1 \\
 & 0.4 & 0.0432 & 0.0503 & 0.0588 & 1 & 1 & 1 \\
 & 0.7 & 0.2150 & 0.2482 & 0.2697 & 1 & 1 & 1 \\
 & 1.0 & 0.5530 & 0.6590 & 0.7490 & 1 & 1 & 1 \\
\hline
\multirow{7}{*}{40} & 0.05 & 0.0002 & 0.0003 & 0.0004 & 1 & 1 & 0.9983 \\
 & 0.08 & 0.0006 & 0.0008 & 0.0011 & 0.9994 & 1 & 0.9989 \\
 & 0.11 & 0.0011 & 0.0015 & 0.0021 & 0.9996 & 0.9995 & 0.9995 \\
 & 0.14 & 0.0018 & 0.0025 & 0.0035 & 0.9995 & 0.9999 & 0.9999 \\
 & 0.4 & 0.0423 & 0.0507 & 0.0574 & 1 & 1 & 1 \\
 & 0.7 & 0.2113 & 0.2481 & 0.2679 & 1 & 1 & 1 \\
 & 1.0 & 0.5491 & 0.6553 & 0.7400 & 1 & 1 & 1 \\
\end{tabular}
\end{center}
\end{table}

\section{Conclusions}
We have presented a random generator of  User Debate Graph instances, where a parameter $\alpha$ is introduced to control 
the expected bipartite polarization of the instances. As previous results with real instances seemed to indicate that instances with polarization away from neutral (0) were easy to solve on average, we wanted to check if this was still the case when working with instances with a more wide set of polarization values. 

On  the one hand, the results obtained are consistent with the ones obtained with real instances, but show that hard to solve instances are in principle possible, at least for very low polarization values, something that is consistent with the fact that the problem is NP-hard.
On the other hand, the results also show that the verification of polarization with this measurement seems difficult only in unrealistic cases, so that it can be used  to monitor polarization in online debates with the final goal to inform solutions for creating healthier social media environments.

Despite the existence of hard instances, they are tightly concentrated around a very 
thin region with very low polarization (that we could argue that it represents a zone of uncommon real instances), and in any case the results about the quality of the solutions obtained with the efficient local search approach 
indicates that it is a good approach to make the computation of the polarization feasible.
Of course, it could be that some specific characteristics of instances coming from some social networks could be not significantly present in our random generator model, or that other measures of polarization could present a different behavior to the one of the measure used in this work. 
So, as future work, we could consider the validation of our model with respect to other social networks and consider alternative measures for the polarization that could make sense in some settings, as the polarization metric introduced in~\cite{DBLP:conf/icwsm/GuerraMCK13} to analyze opinions expressed on the gun control issue in Twitter.

\section*{Acknowledgements}
This work was partially funded by Spanish Project PID2019-111544GB-C22/AEI/10.13039/501100011033 (MICIN), consolideted research group grant 2021-SGR-01615.


\bibliography{bibliografia}
\end{document}